\documentclass[10pt]{article} 
\usepackage[preprint]{rlc}

\usepackage{amssymb}            
\usepackage{mathtools}          
\usepackage{mathrsfs}           
\mathtoolsset{showonlyrefs}     
\usepackage{graphicx}           
\usepackage{subcaption}         
\usepackage[space]{grffile}     
\usepackage{url}                

\chardef\usethreeparttable=0
\chardef\usemathcommands=1
\chardef\usealgorithmicx=0
\chardef\usexcolor=0

\newcommand{\importifndef}[2]{\ifcsname#1\endcsname\else\usepackage{#2}\fi}
\newcommand{\tryusepackage}[1]{\IfPackageLoadedTF{#1}{}{\usepackage{#1}}}
\makeatletter
\newcommand{\newcommandifndef}[3][]{%
  \expandafter\ifcsname#2\endcsname
  \else
    \expandafter\newcommand\csname#2\endcsname[#1]{#3}%
  \fi
}
\makeatother

\if\the\usexcolor
\usepackage[dvipsnames, svgnames, x11names, table]{xcolor}
\fi

\importifndef{algorithm}{algorithm}
\usepackage[noend]{algpseudocode}

\if\the\usealgorithmicx1
\tryusepackage{algorithmicx}
\fi

\tryusepackage{amsmath}
\tryusepackage{amsfonts}
\importifndef{proof}{amsthm}
\tryusepackage{bm}
\tryusepackage{nicefrac}

\tryusepackage{microtype}
\tryusepackage{wrapfig}
\tryusepackage{graphicx}
\tryusepackage{subcaption}

\importifndef{strip}{cuted}

\tryusepackage{array}
\tryusepackage{diagbox}
\tryusepackage{multirow}
\tryusepackage{multicol}
\tryusepackage{booktabs}

\if\the\usethreeparttable1
\tryusepackage{threeparttable}
\fi

\usepackage{xspace}
\usepackage[acronym]{glossaries}
\makeglossaries

\usepackage{rotating}

\tryusepackage{tikz}
\tryusepackage{pgfplots}
\usetikzlibrary{plotmarks}
\usetikzlibrary{positioning}
\pgfplotsset{compat=1.16}

\definecolor{wong1}{rgb}{0.9019607843137255, 0.6235294117647059, 0.0}
\definecolor{wong2}{rgb}{0.33725490196078434, 0.7058823529411765, 0.9137254901960784}
\definecolor{wong3}{rgb}{0.0, 0.6196078431372549, 0.45098039215686275}
\definecolor{wong4}{rgb}{0.9411764705882353, 0.8941176470588236, 0.25882352941176473}
\definecolor{wong5}{rgb}{0.0, 0.4470588235294118, 0.6980392156862745}
\definecolor{wong6}{rgb}{0.8352941176470589, 0.3686274509803922, 0.0}
\definecolor{wong7}{rgb}{0.8, 0.4745098039215686, 0.6549019607843137}

\pgfplotscreateplotcyclelist{cvl}{
solid, wong1, every mark/.append style={solid,fill=\pgfplotsmarklistfill}, mark=*\\
solid, wong2, every mark/.append style={solid,fill=\pgfplotsmarklistfill}, mark=square*\\
solid, wong3, every mark/.append style={solid,fill=\pgfplotsmarklistfill}, mark=triangle*\\
}

\tryusepackage{listings}

\definecolor{codegreen}{rgb}{0,0.6,0}
\definecolor{codegray}{rgb}{0.5,0.5,0.5}
\definecolor{codepurple}{rgb}{0.58,0,0.82}
\definecolor{backcolour}{rgb}{0.95,0.95,0.92}
\definecolor{dkgreen}{rgb}{0,0.6,0}
\definecolor{gray}{rgb}{0.5,0.5,0.5}
\definecolor{mauve}{rgb}{0.58,0,0.82}
 
\lstdefinestyle{lstStyleCode}{
    backgroundcolor=\color{backcolour},
    basicstyle=\ttfamily\footnotesize,
    breakatwhitespace=false,
    captionpos=b,
    keepspaces=true,
    numbers=left,
    numbersep=5pt,
    frame=single,
    showspaces=false,
    showstringspaces=false,
    showtabs=false,
    tabsize=2,
    xleftmargin=2em,
    xrightmargin=2em,
    breaklines=true,
    breakindent=2em,
    framexleftmargin=2em,
    framexrightmargin=2em,
    aboveskip=1em,
    columns=flexible,
    numberstyle=\tiny\color{gray},
    keywordstyle=\color{blue},
    commentstyle=\color{dkgreen},
    stringstyle=\color{mauve},
    breaklines=true,
}

\lstnewenvironment{codeblock}[1][]
    {\lstset{style=lstStyleCode,#1}}{}

\usepackage{tcolorbox}
\newtcbox{\code}{on line, boxrule=0pt, boxsep=0pt, top=2pt, left=2pt, bottom=2pt, right=2pt, colback=gray!15, colframe=white, fontupper={\ttfamily}}

\importifndef{qty}{siunitx}

\importifndef{tablefootnote}{tablefootnote}
\usepackage{colortbl}

\importifndef{labelindent}{enumitem}
\tryusepackage{makecell}

\newcommandifndef[2]{stdmean}{\ensuremath{#2\pm#1}}

\newcommandifndef[2]{bpm}{\textbf{#1} \ensuremath{\pm} \textbf{#2}}
\newcommandifndef[2]{npm}{#1 \ensuremath{\pm} #2}

\newcommandifndef[2]{mbpm}{\ensuremath{\bm{#1} \pm \bm{#2}}}
\newcommandifndef[2]{mppm}{\ensuremath{#1 \pm #2}}
\newcommandifndef[2]{mnpm}{\ensuremath{#1 \pm #2}}

\newcommandifndef[1]{cmt}{{\footnotesize\textcolor{red}{#1}}}
\newcommandifndef[1]{lcmt}{{\footnotesize\textcolor{red}{#1}}\\}
\newcommandifndef[1]{notsure}{\sethlcolor{yellow}\hl{#1}}
\newcommandifndef[1]{todo}{\textcolor{blue}{[TODO: #1]}}
\newcommandifndef[1]{alt}{\sethlcolor{green}\hl{#1}}
\newcommandifndef[1]{fixed}{\iffalse fixed: #1\fi}
\newcommandifndef[1]{rej}{\iffalse#1\fi}
\newcommandifndef[0]{review}{\sethlcolor{yellow}\hl{pls. review:}}
\newcommandifndef[2]{sub}{#2}
\newcommandifndef[2]{car}{#1}

\newcommandifndef[1]{minisection}{\noindent\textbf{#1.~}}

\newcommandifndef[0]{eg}{e.g.\ }
\newcommandifndef[0]{ie}{i.e.\ }
\newcommandifndef[0]{etal}{et al.\ }
\newcommandifndef[0]{cf}{cf.}

\newcommandifndef[2]{Eleft}{\operatorname{\mathbb{E}}_{#1}\left[#2\right.}
\newcommandifndef[2]{Eright}{\left.#2\right.}
\newcommandifndef[0]{EEE}{\mathbb{E}}

\newcommandifndef[0]{density}{p}
\newcommandifndef[0]{proposal}{q}  
\newcommandifndef[0]{target}{p}  
\newcommandifndef[0]{prop}{P}
\newcommandifndef[2]{kl}{\mathrm{D_{KL}}\left(#1\;\middle\|\;#2\right)}
\newcommandifndef[0]{entropy}{\mathcal{H}}
\newcommandifndef[0]{ent}{\mathcal{H}}
\newcommandifndef[0]{tr}{\mathrm{tr}}

\newcommandifndef[0]{ones}{\boldsymbol{1}}
\newcommandifndef[0]{eye}{\boldsymbol{I}}
\newcommandifndef[0]{zeros}{\boldsymbol{0}}

\newcommandifndef[0]{voidarg}{{\,\cdot\,}}

\if\the\usemathcommands1

\newcommandifndef[0]{figleft}{{\em (Left)}}
\newcommandifndef[0]{figcenter}{{\em (Center)}}
\newcommandifndef[0]{figright}{{\em (Right)}}
\newcommandifndef[0]{figtop}{{\em (Top)}}
\newcommandifndef[0]{figbottom}{{\em (Bottom)}}
\newcommandifndef[0]{captiona}{{\em (a)}}
\newcommandifndef[0]{captionb}{{\em (b)}}
\newcommandifndef[0]{captionc}{{\em (c)}}
\newcommandifndef[0]{captiond}{{\em (d)}}

\newcommandifndef[1]{newterm}{{\bf #1}}

\def\eqref#1{Eq.~(\ref{#1})}

\def\1{\bm{1}}
\newcommandifndef[0]{train}{\mathcal{D}}
\newcommandifndef[0]{valid}{\mathcal{D_{\mathrm{valid}}}}
\newcommandifndef[0]{test}{\mathcal{D_{\mathrm{test}}}}

\DeclareMathAlphabet{\mathsfit}{\encodingdefault}{\sfdefault}{m}{sl}
\SetMathAlphabet{\mathsfit}{bold}{\encodingdefault}{\sfdefault}{bx}{n}
\newcommandifndef[1]{tens}{\bm{\mathsfit{#1}}}

\def\gR{{\mathcal{R}}}

\newcommandifndef[1]{etens}{\mathsfit{#1}}

\newcommandifndef[0]{pdata}{p_{\rm{data}}}
\newcommandifndef[0]{ptrain}{\hat{p}_{\rm{data}}}
\newcommandifndef[0]{Ptrain}{\hat{P}_{\rm{data}}}
\newcommandifndef[0]{pmodel}{p_{\rm{model}}}
\newcommandifndef[0]{Pmodel}{P_{\rm{model}}}
\newcommandifndef[0]{ptildemodel}{\tilde{p}_{\rm{model}}}
\newcommandifndef[0]{pencode}{p_{\rm{encoder}}}
\newcommandifndef[0]{pdecode}{p_{\rm{decoder}}}
\newcommandifndef[0]{precons}{p_{\rm{reconstruct}}}

\newcommandifndef[0]{laplace}{\mathrm{Laplace}} 

\newcommandifndef[0]{E}{\mathbb{E}}
\newcommandifndef[0]{Ls}{\mathcal{L}}
\newcommandifndef[0]{R}{\mathbb{R}}
\newcommandifndef[0]{emp}{\tilde{p}}
\newcommandifndef[0]{lr}{\alpha}
\newcommandifndef[0]{reg}{\lambda}
\newcommandifndef[0]{rect}{\mathrm{rectifier}}
\newcommandifndef[0]{softmax}{\mathrm{softmax}}
\newcommandifndef[0]{sigmoid}{\sigma}
\newcommandifndef[0]{softplus}{\zeta}
\newcommandifndef[0]{KL}{D_{\mathrm{KL}}}
\newcommandifndef[0]{Var}{\mathrm{Var}}
\newcommandifndef[0]{standarderror}{\mathrm{SE}}
\newcommandifndef[0]{Cov}{\mathrm{Cov}}
\newcommandifndef[0]{normlzero}{L^0}
\newcommandifndef[0]{normlone}{L^1}
\newcommandifndef[0]{normltwo}{L^2}
\newcommandifndef[0]{normlp}{L^p}
\newcommandifndef[0]{normmax}{L^\infty}

\newcommandifndef[0]{parents}{Pa} 

\newcommandifndef[2]{parfrac}{\frac{\partial{#1}}{\partial{#2}}}
\newcommandifndef[3]{chainrule}{\parfrac{#1}{#2}\cdot\parfrac{#2}{#3}}

\fi

\usepackage{enumitem}
\def\ours{UniCon\xspace}

\title{UniCon: A Unified System for Efficient Robot Learning Transfers}

\author{Yunfeng Lin \\
    linyunfeng@sjtu.edu.cn \\
    Shanghai Jiao Tong University \\
    Shanghai Artificial Intelligence Laboratory \\
    Shanghai, China \\
    \And
    Li Xu \\
    lixu@changan.com.cn \\
    AI Laboratory, \\
    Chongqing Changan Automobile Co. Ltd. \\
    Chongqing, China \\
    \And
    Yong Yu \\
    yyu@apex.sjtu.edu.cn \\
    Shanghai Jiao Tong University \\
    Shanghai, China \\
    \And
    Jiangmiao Pang \\
    pangjiangmiao@pjlab.org.cn \\
    Shanghai Artificial Intelligence Laboratory \\
    Shanghai, China \\
    \And
    Weinan Zhang \thanks{Corresponding author.} \\
    wnzhang@sjtu.edu.cn \\
    Shanghai Jiao Tong University \\
    Shanghai Artificial Intelligence Laboratory \\
    Shanghai, China }

\begin{document}

\maketitle

\begin{figure*}[h]
\centering
\includegraphics[width=1.0\textwidth]{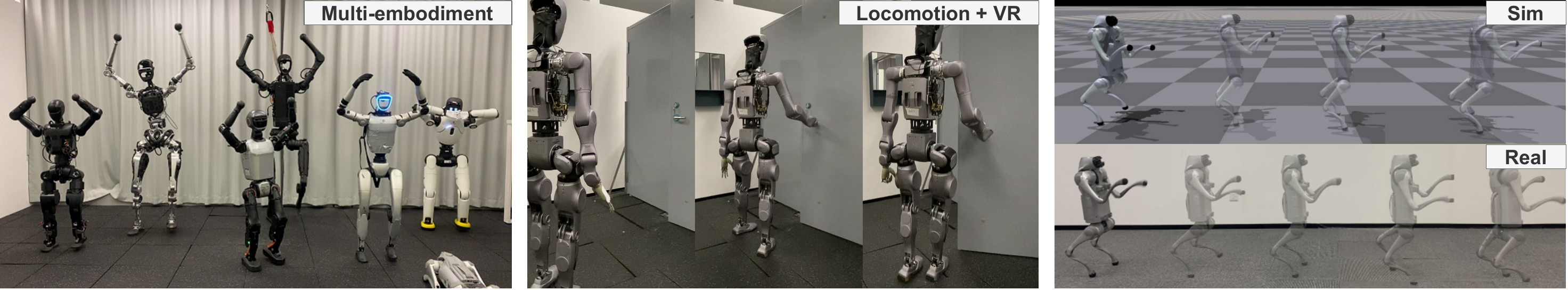}
\captionof{figure}{
Representative use cases of \ours: 
\textbf{Left:} synchronized and reusable locomotion across heterogeneous robots. 
\textbf{Middle:} Modular interoperation of RL policies with VR teleoperation. 
\textbf{Right:} Real‑to‑sim data recording and analysis for diagnosing transfer gaps. 
Data and control flow are standardized across platforms, reducing integration effort and improving efficiency.
}
\end{figure*}


\begin{abstract}
Deploying learning-based controllers across heterogeneous robots is challenging due to platform differences, inconsistent interfaces, and inefficient middleware.
To address these issues, we present \ours, a lightweight framework that standardizes states, control flow, and instrumentation across platforms.
It decomposes workflows into execution graphs with reusable components, separating system states from control logic to enable plug-and-play deployment across various robot morphologies.
Unlike traditional middleware, it prioritizes efficiency through batched, vectorized data flow, minimizing communication overhead and improving inference latency. 
This modular, data-oriented approach enables seamless sim-to-real transfer with minimal re-engineering.
We demonstrate that \ours reduces code redundancy when transferring workflows and achieves higher inference efficiency compared to ROS-based systems.
Deployed on over 12 robot models from 7 manufacturers, it has been successfully integrated into ongoing research projects, proving its effectiveness in real-world scenarios.
\end{abstract}

\section{Introduction}

Deploying learning‑based controllers across heterogeneous robots remains costly due to platform differences and tightly coupled interfaces.
These incompatibilities force developers to repeatedly re-implement the same functionalities for each combination of algorithms, robots and simulators.
While domain randomization and related sim-to-real techniques \citep{randomization1,randomization2} shrink the dynamics gap, they do not simplify this last-mile integration.
Widely used middleware such as ROS and ROS~2~\citep{ros,ros2} attempt to provide a standardized communication layer between customized components, but their general‑purpose design introduces \textbf{latency and overhead} that hinder high‑frequency policy inference~\citep{pytorch,Ansel2024PyTorch2F,onnx}.
At the same time, robot manufacturers complicate the ecosystem further by shipping \textbf{inconsistent interfaces}, making integration and transferability difficult.


To address these challenges, we present \ours, a lightweight framework for robot learning deployments that provides the unifying infrastructure needed to standardize states, control flow, and instrumentation across platforms.
\ours decomposes workflows into execution graphs with reusable basic units, and separates system states -- such as proprioceptive, exteroceptive, and control signals -- from the logic that consumes them.
This enables \textbf{plug‑and‑play deployment} across diverse morphologies including quadrupeds, humanoids, and manipulators, while unifying hardware with simulators for seamless transfer.


Compared to ROS, our architecture adopts a data-oriented modular paradigm with separation between runtime states and control logic.
To maximize efficiency, we favor single-threaded computing~\citep{ros2compose} over multiprocessing, and vectorized data flow over message passing.
On the other hand, interoperability with existing software is still guaranteed by extending the underlying states storage to multiple communication backends~\citep{zeromq2010,ddsfoundation2025}.
We show that this reduces code redundancy and improves inference latency in our cross-embodiment experiments.

Our contributions are threefold:
\begin{itemize}
    \item \textbf{Unified control framework:} We introduce \ours, a lightweight and efficient framework that bridges mainstream simulators and physical hardware for robot learning.
    \item \textbf{Modular, data-oriented design:} We demonstrate that \ours enables efficient sim-to-real transfer across diverse robot embodiments with little to no re-engineering effort.
    \item \textbf{Improved inference efficiency:} We show that \ours achieves higher inference efficiency compared to ROS-based and ad-hoc framework stacks by reducing interfacing and communication overhead.
\end{itemize}

\ours has already been deployed on over \textbf{12 robot models} spanning \textbf{7 manufacturers}, and has supported \textbf{3 research projects} in the Shanghai Artificial Intelligence Laboratory, demonstrating its versatility in real research scenarios.
We release our framework as open-source at \url{https://github.com/creeperlin/unicon}.

\section{Framework}

A key obstacle in sim‑to‑real and cross-embodiment transfer is the tight coupling of system data flows with customized control logic.
Middleware such as ROS enforces a message passing pattern but incurs latency, making it unsuitable for learning‑based controllers.
Our design instead adopts a modular, data‑oriented paradigm that cleanly separates states from logic, emphasizing reusability and efficiency.
This approach contrasts with object‑oriented programming practices~\citep{10.5555/1407387}, where data and routines are encapsulated together, and instead aligns more closely with the Entity Component System architecture widely used in the game industry~\citep{overwatch2017gdc} and modern simulation engines~\citep{shacklett23madrona}.

\begin{figure}[htb]
  \centering
  \includegraphics[width=0.6\linewidth]{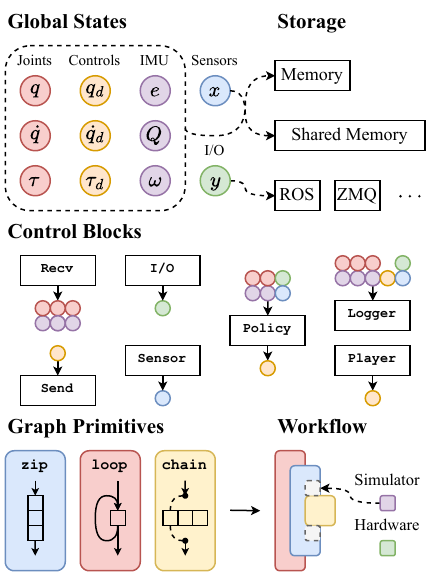}
  \caption{
Architecture of \ours: (a) global system states with switchable storage backends; (b) modular control blocks covering platform and inference; (c) control flow graph primitives for workflow composition; and (d) unified integration with simulators and hardware.
}
  \label{fig:overview}
\end{figure}

\subsection{Vectorized States}
We formulate all runtime states of a robot workflow as global array objects of numerical data types under different labels.
This comes from the fact that the data passed in structured messages often takes a fixed format for each channel if the hardware participants remains constant.
Therefore, rather than transmitting full self‑describing objects, only the essential multidimensional data are exchanged, eliminating unnecessary encapsulation and aligning with contiguous memory layouts favored by simulators and hardware interfaces.

Typically, the states include robot sensory inputs such as motor position and velocities, as well as orientation and angular velocity reported by the IMU.
Similarly, motor commands such as position and torque controls are also states which are written to by controllers and read from by hardware interfaces.
In the case where state definitions of custom components differ from the actual hardware, transparent mappings with efficient vectorized indexing operations are inserted to align properties such as joint order.

\subsection{Modularized Logic}
Following the vectorized formulation, we break down workflows into reusable Control Blocks (CB) bridged by global states.
For example, operations like receiving hardware states, executing policy inference, handling user inputs, and sending hardware commands are split into individual blocks and connected into an execution graph through standardized control flow.
This grants users the ability to shift between controllers, policies and hardware by switching between CBs, without changing the rest of the graph.

Formally, we assume the global state space $S=\{s_a,s_b,\dots,s_n\}, s_i\in \gR^{k_i}$ is the common domain of all operating logic.
Then we define a CB as a pure function with regard to $S$.
It can read a subset of $S$ as inputs and outputs updated values of some other elements in $S$, and a boolean $r$ indicating its termination.

\begin{equation}
f: S_{in} \mapsto S_{out}\times \{0,1\}, S_{in}, S_{out} \subseteq S.
\end{equation}

Stepping the function results in a sequence of calls:

\begin{equation}
T_f = (f_1, f_2, \dots, f_n).
\end{equation}

A workflow is then constructed by nesting the CBs, forming an execution graph.
We provide basic control flow components in similar fashion to functional programming~\citep{Hughes1989WhyFP} for composing customized workflows.
For example, \code{loop} terminates the inner block when the predicate is satisfied, \code{zip} executes multiple blocks in sequence, and \code{chain} exhausts nested blocks in order until they all halt.

\begin{equation}
\begin{aligned}
  \operatorname{loop}(T_g, p) &= (g_1, g_2, \dots, g_k), p(k) = 1, \\
  \operatorname{zip}(T_g, T_h) &= ((g_1, h_1), (g_2, h_2), \dots, (g_n, h_n)), \\
  \operatorname{chain}(T_g, T_h) &= (g_1, \dots, g_m, h_1, \dots, h_n).
\end{aligned}
\end{equation}

This enables configurations via text formats and possibly GUIs, while also leaving directives for code generation.

\subsection{Software Interoperability}

\minisection{Communications}
\ours connects its state storage backend to external clients with consistent semantics: writing to arrays publishes data, reading retrieves the latest values.
Multiple backends provide flexible performance: lock‑free local or shared memory for zero‑copy access~\citep{Poehnl2019Iceoryx}, message queues for distributed setups~\citep{zeromq2010}, and ROS bridges for legacy integration.

\minisection{Platform support}
Simulation and hardware platforms are integrated uniformly as state readers and writers.
Each platform adapter provides three callable blocks: \code{recv} for updating states, \code{send} for sending controls, and \code{close} for teardown, while automatically aligning parameters such as PID gains and DoF limits to ensure consistent control semantics~\citep{Chitta2017roscontrolAG}.
Supported platforms are summarized in \tableautorefname~\ref{tab:platforms}.

\minisection{Peripherals}
We also incorporate frequently used peripheral tools and devices as plug-and-play CBs.
Input devices such as joysticks~\citep{linuxinput2025} and VR~\citep{vuer} are supported with multiple implementations for coverage,
while sensors including RGBD~\citep{realsense} cameras and LiDARs are connected either through platform proxy or standalone clients; high-bandwidth image data are handled through their native interfaces and referenced via lightweight handles to preserve efficiency.
Visualizers such as Foxglove and Meshcat~\citep{qin2023anyteleop} are also ready to be wired to selected states.

\subsection{Programming Interface}

\ours exposes states as typed NumPy~\citep{harris2020array} arrays, enabling users to perform vectorized operations in Python.
Workflows can be defined either programmatically or textually, with user logic seamlessly integrated via parameter-less boolean functions following our CB design.
This causes minimal intrusion while allowing users to fully customize their control flow as needed.

\subsection{Workflow Transfers}

\ours enables seamless transfer between three workflows: \textbf{Sim-to-Sim} validation, \textbf{Sim-to-Real} deployment, and \textbf{Real-to-Sim} instrumentation, all operating on a unified data abstraction.
Sim-to-Sim and Sim-to-Real are trivial since the system CBs remain interchangeable within the execution graph.
For real deployments, additional safety blocks can be inserted to handle startup, shutdown, and runtime monitoring without altering the rest of the graph.

Real-to-Sim uses recorder and replay blocks on physical states and built‑in metrics to quantify the reality gap.
For open‑loop and repeatable rollouts, mean squared error (MSE) performs step‑wise comparison since inputs are exactly replayed.
For closed‑loop controls like locomotion, we propose to unfold the MSE loss to prevent temporal misalignments:

\begin{equation}
\mathcal{L}= \frac{1}{n}\min_j \sum_{i=0}^{n} \|T_{i+j} - \hat{T}_i\|_2 + \frac{1}{n}\min_j \sum_{i=0}^{n} \|T_{i} - \hat{T}_{i+j}\|_2,
\end{equation}

\noindent where $T$ represents the real robot trajectory, $\hat{T}$ represents the simulated trajectory, and $j$ is the temporal shift minimized to find the best alignment between the two sequences.



\section{Evaluation}

\begin{table}[!t]
  \centering
  \caption{Supported simulators and hardware platforms in \ours.}
  \label{tab:platforms}
  {\small
  \begin{tabular}{l l l}
    \toprule
    \textbf{Simulators} & \textbf{Humanoids} & \textbf{Quadrupeds} \\
    \midrule
    IsaacGym~\citep{isaacgym}   & Unitree H1~\citep{unitree}   & Unitree A1 \\
    IsaacSim~\citep{nvidia2023isaacsim}   & Unitree  G1   & Unitree Aliengo \\
    IsaacLab~\citep{orbit}   & Unitree H1-2 & Unitree Go2 \\
    MuJoCo~\citep{mujoco} & Fourier GR1~\citep{fftai} &  \\
    \cmidrule{3-3}
    MuJoCo MJX~\citep{mujoco_playground_2025} & Fourier N1 & \textbf{Manipulators} \\
    \cmidrule{3-3}
    PyBullet~\citep{pybullet}   & AgiBot X2~\citep{agibot2023x2}  & ARX~\citep{arxroboticsx2025github} \\
    Webots~\citep{Michel2004CyberboticsLW} & Dobot Atom~\citep{atom} & ROHand~\citep{rohand2025oymotion} \\
    Genesis~\citep{Genesis}     & PND Adam~\citep{pnd} & Unitree Dex3-1\\
    Gazebo~\citep{Koenig2004DesignAU}     & AzureLoong~\citep{oghr2} & Inspire RH56 \\
    Newton~\citep{newton2025github}       & Booster T1~\citep{t1} & XHand1 \\
    \bottomrule
  \end{tabular}
  }
\end{table}

We evaluate our framework on deployment tasks across quadrupeds, humanoids, and manipulators,
focusing on transfer effort, inference efficiency and workflow usability.

\subsection{Transfer Effort}

We target the locomotion task for humanoids and quadrupeds where policies with the same training paradigm are deployed on multiple platforms~\citep{xue2025unified}.
Without a framework, the inference often needs to be re‑implemented due to vendor‑specific stacks.
We report the Source Lines of Code (SLOC) required to transfer the original workflow on Unitree H1 to the MuJoCo simulator and other robots.
As shown in \tableautorefname~\ref{tab:transfer}, the Ad‑hoc code needs substantial efforts to transfer, especially for the PND Adam model with its completely different APIs, while \ours remains lightweight and requires no extra effort when switching deployment targets.


\begin{table}[htb]
  \centering
\setlength{\extrarowheight}{0pt}
\addtolength{\extrarowheight}{\aboverulesep}
\addtolength{\extrarowheight}{\belowrulesep}
\setlength{\aboverulesep}{0pt}
\setlength{\belowrulesep}{0pt}
  \caption{Comparison of transfer effort across workflows (code changes, in SLOC).}
  \label{tab:transfer}
  \begin{tabular}{l c c c c c c c c c}
    \toprule
    \multirow{2}{*}{Framework} & \multicolumn{5}{c}{Components} & \multicolumn{4}{c}{Inference workflows} \\
    \cmidrule(lr){2-6} \cmidrule(lr){7-10}
                               & H1 & Sim & G1 & Adam & NN & H1 & Sim & G1 & Adam \\
    \midrule
    Ad-hoc & \multicolumn{5}{c}{852 (original code base for H1)} & 0 &  507 & 627 & 770 \\
    \ours (ours)  & 393 & 382 & 35 & 394 & 350 & 0 & 0 & 0 & 0 \\
    \bottomrule
  \end{tabular}
\end{table}








\subsection{Framework Efficiency}

We measure the framework efficiency in the form of operation latency overhead.
The evaluation runs on the same H1 model with identity joint position control (\(q_{d}=q\)) at 50Hz in place of actual inference. 
For this robot, system states are refreshed at around 500Hz by the SDK~\citep{unitreeSDK2python2025}.
The synchronous (Sync) setup queries the global buffer directly at each control cycle, while the asynchronous (Async) setup uses a callback to receive the latest states.
In the table, \emph{Recv} denotes the time to obtain the latest system states, \emph{Send} denotes the time to transmit the resulting control signal, and \emph{End-to-end} measures the latency between sending a control signal and the timestamp of the dependent states.
Because the state timestamp originates from a monotonic clock on another onboard system, we manually align clocks with an inferred time origin for this row.

Results in \tableautorefname~\ref{tab:latency} show that \ours introduces near‑zero overhead compared to Python SDK implementations and significantly outperforms ROS.
This efficiency arises from binding global state buffers into the data reader/writer and compiling~\citep{pybind11github,RoboJuDo} alongside the SDK to eliminate redundant data copies.
During actual inference, control blocks can be arranged so that the critical path contains only the data producers required by the policy, deferring non‑critical communication and synchronization.

\begin{table}[htb]
  \centering

  \caption{Comparison of framework efficiency on H1 (operation latency, in $\mu$s).}
  \label{tab:latency}
  \begin{tabular}{lcccc}
    \toprule
    \multirow{2}{*}{Operation} & \multicolumn{4}{c}{Frameworks} \\
    \cmidrule(lr){2-5}
              & SDK (Sync) & SDK (Async) & ROS 2 & \ours (ours) \\
    \midrule
    Recv   & 1261 $\pm$ 150 & - & \textbf{50 $\pm$ 88} & \textbf{63 $\pm$ 15} \\
    Send      & 1011 $\pm$ 72  & 1001 $\pm$ 65  & 587 $\pm$ 179        & \textbf{190 $\pm$ 45} \\
    End-to-end\footnotemark[1] & 1805 $\pm$ 1559  &  1447 $\pm$ 936  & 1658 $\pm$ 882      & \textbf{732 $\pm$ 749} \\
    \bottomrule
    \multicolumn{5}{l}{\scriptsize{\footnotemark[1] Time between send and state timestamp, applying an inferred offset due to relative system clocks.}} \\
  \end{tabular}
\end{table}



\subsection{Workflow Usability}

While ROS is the go-to choice for building complex robotic applications, \ours offers comparable usability with built-in connection to mainstream sensors, input devices, motion libraries and messaging layers including ROS itself.
We demonstrate this by extending the inference pipeline to wholebody control, incorporating inverse kinematics~\citep{Buss2004IntroductionTI,pink}, dexterous hand retargeting~\citep{qin2023anyteleop}, and VR teleoperation~\citep{Cheng2024OpenTeleVisionTW}, all modularized and distributed across onboard and LAN workstations.

\begin{figure}[htb]
  \centering
  \resizebox{0.9\textwidth}{!}{%
    \begin{tabular}{ccc}
      \includegraphics[width=0.4090\linewidth]{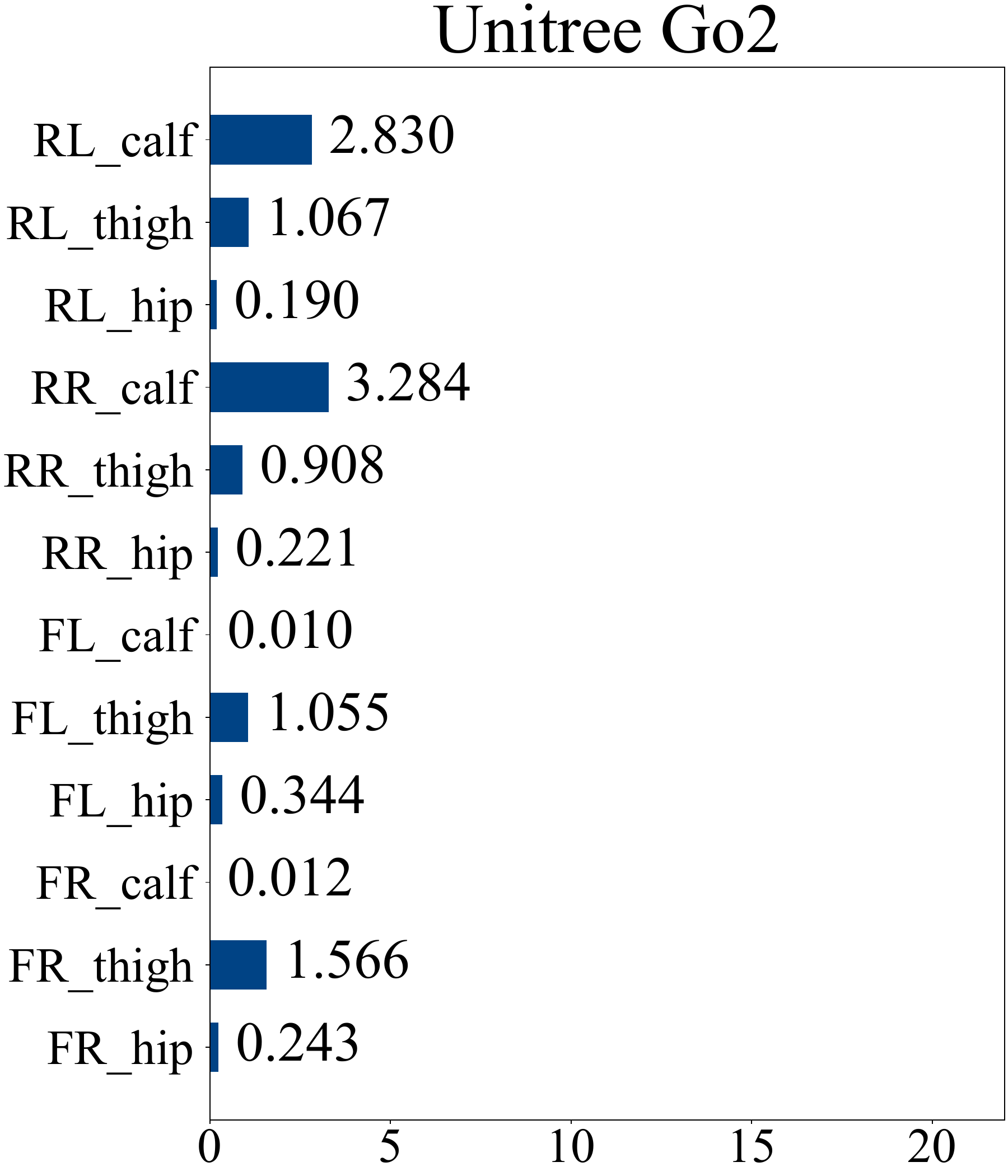} &
      \includegraphics[width=0.4090\linewidth]{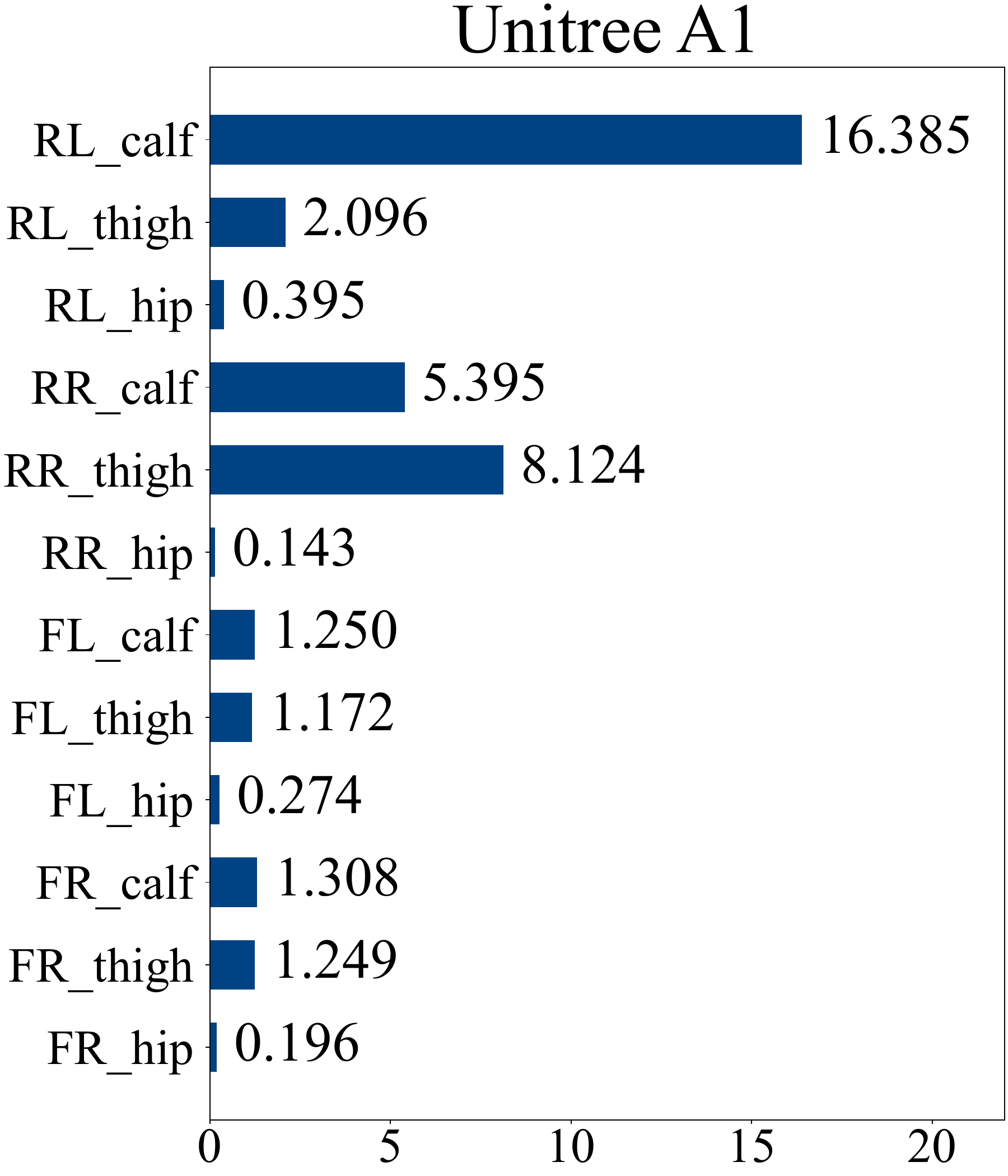} &
      \includegraphics[width=0.1515\linewidth]{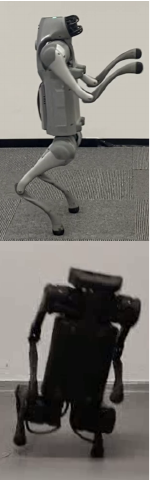}
    \end{tabular}
  }
  \caption{
Real-to-sim analysis of inference trajectories. 
\textbf{Left \& Middle:} Reality gap quantified per joint position using built-in metrics, showing deviations in the A1's rear left calf joint. 
\textbf{Right:} Stable Go2 standing (top) and A1 just before fallover (bottom).
}
  \label{fig:realtosim}
\end{figure}

We further showcase real‑to‑sim analysis by comparing recorded and replayed trajectories across systems.
Using the built‑in analyzer, discrepancies can be measured element‑wise and frame‑wise to identify sources of the reality gap.
As shown in \figureautorefname~\ref{fig:realtosim}, a bipedal policy stable in the IsaacGym simulation but failing on the A1 robot exhibits significant deviation in the rear left calf joint, leading to overheating and eventual loss of balance.
In contrast, the Go2 model shows a smaller gap under similar settings.

\section{Related Work}


\subsection{Learning-based Robot Controllers}

Learning-based controllers, particularly reinforcement learning (RL), have enabled locomotion across quadrupeds~\citep{lee2020learning, smith2022, rma, hoeller2024anymal}, wheeled robots~\citep{lee2024learning}, and humanoids~\citep{gu2024advancing,xue2025unified,wang2025beamdojo,huang2025learning}, achieving agile gaits~\citep{amp-wu,he2024agile}, terrain adaptation~\citep{tert,parkour1,parkour2,Lai2024WorldMP}, whole-body control~\citep{cheng2024express,ji2024exbody2}, and motion tracking~\citep{amp,yin2025unitrackerlearninguniversalwholebody}.
These advances highlight the promise of RL for scalable locomotion, but also the challenge of sim-to-real transfer: policies trained in simulation often fail to generalize due to discrepancies in dynamics and perception~\citep{comparesim,da2025survey}.

To address this, domain randomization~\citep{randomization1,OpenAI2019SolvingRC,randomization2,campanaro2022} and adaptation~\citep{minimizing} improve robustness, while real-to-sim system identification (SysId) methods calibrate simulators with real-world data~\citep{dong2024easi,he2025asap,xu2025deal}.
UniCon complements these strategies by introducing a normalization pipeline and hardware-agnostic interface abstraction, enabling cross-platform inference, data gathering and analysis.

\subsection{Robot Application Middleware}

Robot application middleware provides critical abstractions and communication infrastructure for integrating perception, planning, and control components across heterogeneous robotic systems~\citep{Gerkey2003ThePP,Metta2006YARPYA,Bruyninckx2001OpenRC,Utz2002MiroM}.
Frameworks such as the Robot Operating System (ROS) have become de facto standards in academia and industry due to their modularity, device driver ecosystem, and support for distributed computing~\citep{ros,ros2,Chitta2012MoveItT}.
Similarly, PX4~\citep{Meier2015PX4AN} offers a full-stack flight control system for aerial robotics, facilitating high-performance real-time control with rich simulation integration.

While these middleware platforms simplify software development and system orchestration, they generally lack standardized abstractions for cross-platform strategy deployment or sim-to-real policy learning.
UniCon complements these frameworks by introducing a domain-aware control layer specifically tailored for learning-based policy modularization and transferability. Rather than replacing existing middleware, UniCon operates orthogonally -- interfacing with existing environments to unify and reuse control logic across embodiments.

\section{Conclusion}

\ours offers a unified control layer in robot learning deployments by separating vectorized global states from modular control blocks, greatly reducing hardware transfer effort and enabling low-latency inference.
Its successful use across robots and research projects shows strong potential for broader community adoption.
Future work includes multi‑language APIs and code generation to broaden platform coverage.

\subsubsection*{Acknowledgments}
The SJTU team is supported by National Natural Science Foundation of China (62322603), Shanghai Municipal Science and Technology Major Project (2025SHZDZX025D08) and Shanghai Artificial Intelligence Laboratory.
We also thank Fourier Intelligence and ByteDance Seed for their early support with hardware and experiments.

\bibliography{bibs/bib3,bibs/bib0}

@article{rma,
  title={Rma: Rapid motor adaptation for legged robots},
  author={Kumar, Ashish and Fu, Zipeng and Pathak, Deepak and Malik, Jitendra},
  journal={arXiv preprint arXiv:2107.04034},
  year={2021}
}

@inproceedings{randomization1,
  title={Sim-to-real transfer of robotic control with dynamics randomization},
  author={Peng, Xue Bin and Andrychowicz, Marcin and Zaremba, Wojciech and Abbeel, Pieter},
  booktitle={2018 IEEE international conference on robotics and automation (ICRA)},
  pages={3803--3810},
  year={2018},
  organization={IEEE}
}

@inproceedings{randomization2,
  title={Dynamics randomization revisited: A case study for quadrupedal locomotion},
  author={Xie, Zhaoming and Da, Xingye and van de Panne, Michiel and Babich, Buck and Garg, Animesh},
  booktitle={2021 IEEE International Conference on Robotics and Automation (ICRA)},
  pages={4955--4961},
  year={2021},
  organization={IEEE}
}

@article{minimizing,
  title={Minimizing energy consumption leads to the emergence of gaits in legged robots},
  author={Fu, Zipeng and Kumar, Ashish and Malik, Jitendra and Pathak, Deepak},
  journal={arXiv preprint arXiv:2111.01674},
  year={2021}
}

@Misc{unitree,
  title = {Unitree Robotics},
  author = {Unitree},
  howpublished = {\url{https://www.unitree.com/}},
  year = 2022,
}

@inproceedings{tert, 
address={London, United Kingdom}, 
title={Sim-to-Real Transfer for Quadrupedal Locomotion via Terrain Transformer}, 
rights={https://doi.org/10.15223/policy-029}, 
ISBN={9798350323658}, 
url={https://ieeexplore.ieee.org/document/10160497/}, 
DOI={10.1109/ICRA48891.2023.10160497}, 
booktitle={2023 IEEE International Conference on Robotics and Automation (ICRA)}, 
publisher={IEEE}, 
author={Lai, Hang and Zhang, Weinan and He, Xialin and Yu, Chen and Tian, Zheng and Yu, Yong and Wang, Jun}, 
year={2023}, 
month=may, 
pages={5141–5147}, 
language={en}
}

@inproceedings{amp,
  title={Adversarial motion priors make good substitutes for complex reward functions},
  author={Escontrela, Alejandro and Peng, Xue Bin and Yu, Wenhao and Zhang, Tingnan and Iscen, Atil and Goldberg, Ken and Abbeel, Pieter},
  booktitle={2022 IEEE/RSJ International Conference on Intelligent Robots and Systems (IROS)},
  pages={25--32},
  year={2022},
  organization={IEEE}
}

@article{amp-wu,
  title={Learning robust and agile legged locomotion using adversarial motion priors},
  author={Wu, Jinze and Xin, Guiyang and Qi, Chenkun and Xue, Yufei},
  journal={IEEE Robotics and Automation Letters},
  year={2023},
  publisher={IEEE}
}

@article{parkour1,
  title={Robot parkour learning},
  author={Zhuang, Ziwen and Fu, Zipeng and Wang, Jianren and Atkeson, Christopher and Schwertfeger, Soeren and Finn, Chelsea and Zhao, Hang},
  journal={arXiv preprint arXiv:2309.05665},
  year={2023}
}

@inproceedings{parkour2,
  title={Extreme parkour with legged robots},
  author={Cheng, Xuxin and Shi, Kexin and Agarwal, Ananye and Pathak, Deepak},
  booktitle={2024 IEEE International Conference on Robotics and Automation (ICRA)},
  pages={11443--11450},
  year={2024},
  organization={IEEE}
}

@article{hoeller2024anymal,
  title={Anymal parkour: Learning agile navigation for quadrupedal robots},
  author={Hoeller, David and Rudin, Nikita and Sako, Dhionis and Hutter, Marco},
  journal={Science Robotics},
  volume={9},
  number={88},
  pages={eadi7566},
  year={2024},
  publisher={American Association for the Advancement of Science}
}

@article{lee2024learning,
  title={Learning robust autonomous navigation and locomotion for wheeled-legged robots},
  author={Lee, Joonho and Bjelonic, Marko and Reske, Alexander and Wellhausen, Lorenz and Miki, Takahiro and Hutter, Marco},
  journal={Science Robotics},
  volume={9},
  number={89},
  pages={eadi9641},
  year={2024},
  publisher={American Association for the Advancement of Science}
}

@article{lee2020learning,
  title={Learning quadrupedal locomotion over challenging terrain},
  author={Lee, Joonho and Hwangbo, Jemin and Wellhausen, Lorenz and Koltun, Vladlen and Hutter, Marco},
  journal={Science robotics},
  volume={5},
  number={47},
  pages={eabc5986},
  year={2020},
  publisher={American Association for the Advancement of Science}
}

@article{gu2024advancing,
  title={Advancing humanoid locomotion: Mastering challenging terrains with denoising world model learning},
  author={Gu, Xinyang and Wang, Yen-Jen and Zhu, Xiang and Shi, Chengming and Guo, Yanjiang and Liu, Yichen and Chen, Jianyu},
  journal={arXiv preprint arXiv:2408.14472},
  year={2024}
}

@article{huang2025learning,
  title={Learning Humanoid Standing-up Control across Diverse Postures},
  author={Huang, Tao and Ren, Junli and Wang, Huayi and Wang, Zirui and Ben, Qingwei and Wen, Muning and Chen, Xiao and Li, Jianan and Pang, Jiangmiao},
  journal={arXiv preprint arXiv:2502.08378},
  year={2025}
}

@article{wang2025beamdojo,
  title={Beamdojo: Learning agile humanoid locomotion on sparse footholds},
  author={Wang, Huayi and Wang, Zirui and Ren, Junli and Ben, Qingwei and Huang, Tao and Zhang, Weinan and Pang, Jiangmiao},
  journal={arXiv preprint arXiv:2502.10363},
  year={2025}
}

@inproceedings{Gerkey2003ThePP,
  title={The Player/Stage Project: Tools for Multi-Robot and Distributed Sensor Systems},
  author={Brian P. Gerkey and Richard T. Vaughan and Andrew Howard},
  year={2003},
  url={https://api.semanticscholar.org/CorpusID:806182}
}

@article{Metta2006YARPYA,
  title={YARP: Yet Another Robot Platform},
  author={Giorgio Metta and Paul M. Fitzpatrick and Lorenzo Natale},
  journal={International Journal of Advanced Robotic Systems},
  year={2006},
  volume={3},
  url={https://api.semanticscholar.org/CorpusID:11757034}
}

@article{Bruyninckx2001OpenRC,
  title={Open robot control software: the OROCOS project},
  author={Herman Bruyninckx},
  journal={Proceedings 2001 ICRA. IEEE International Conference on Robotics and Automation (Cat. No.01CH37164)},
  year={2001},
  volume={3},
  pages={2523-2528 vol.3},
  url={https://api.semanticscholar.org/CorpusID:15556265}
}

@article{Utz2002MiroM,
  title={Miro - middleware for mobile robot applications},
  author={Hans Utz and Stefan Sablatn{\"o}g and Stefan Enderle and Gerhard K. Kraetzschmar},
  journal={IEEE Trans. Robotics Autom.},
  year={2002},
  volume={18},
  pages={493-497},
  url={https://api.semanticscholar.org/CorpusID:14188093}
}

@article{Poehnl2019Iceoryx,
  author       = {Michael P{\"o}hnl},
  title        = {Eclipse iceoryx: Enabling Virtually Limitless Data Transmissions at Constant Time},
  journal      = {Eclipse Newsletter},
  year         = {2019},
  month        = {December},
  publisher    = {The Eclipse Foundation},
  url          = {https://www.eclipse.org/community/eclipse_newsletter/2019/december/4.php},
}

@book{10.5555/1407387,
author = {Booch, Grady and Maksimchuk, Robert and Engle, Michael and Young, Bobbi and Conallen, Jim and Houston, Kelli},
title = {Object-oriented analysis and design with applications, third edition},
year = {2007},
isbn = {9780201895513},
publisher = {Addison-Wesley Professional},
edition = {Third}
}

@misc{RoboJuDo,
  author = {Zhuang, Hans and Dsixy and artpli},
  title = {A plug-and-play deploy framework for robots. Just deploy, just do.},
  url = {https://github.com/HansZ8/RoboJuDo},
  year = {2025}
}

@misc{unitreeSDK2python2025,
  author       = {Unitree Robotics Developers},
  title        = {Unitree SDK2 Python},
  year         = {2025},
  howpublished = {\url{https://github.com/unitreerobotics/unitree_sdk2_python}},
  note         = {Accessed: December 15, 2025}
}

@article{he2025asap,
          title={ASAP: Aligning Simulation and Real-World Physics for Learning Agile Humanoid Whole-Body Skills},
          author={He, Tairan and Gao, Jiawei and Xiao, Wenli and Zhang, Yuanhang and Wang, Zi and Wang, Jiashun and Luo, Zhengyi and He, Guanqi and Sobanbabu, Nikhil and Pan, Chaoyi and Yi, Zeji and Qu, Guannan and Kitani, Kris and Hodgins, Jessica and Fan, Linxi "Jim" and Zhu, Yuke and Liu, Changliu and Shi, Guanya},
          journal={arXiv preprint arXiv:2502.01143},
          year={2025}
        }

@article{da2025survey,
  title={A Survey of Sim-to-Real Methods in RL: Progress, Prospects and Challenges with Foundation Models},
  author={Da, Longchao and Turnau, Justin and Kutralingam, Thirulogasankar Pranav and Velasquez, Alvaro and Shakarian, Paulo and Wei, Hua},
  journal={arXiv preprint arXiv:2502.13187},
  year={2025}
}

@inproceedings{
dong2024easi,
title={{EASI}: Evolutionary Adversarial Simulator Identification for Sim-to-Real Transfer},
author={Haoyu Dong and Huiqiao Fu and Wentao Xu and Zhehao Zhou and Chunlin Chen},
booktitle={The Thirty-eighth Annual Conference on Neural Information Processing Systems},
year={2024},
url={https://openreview.net/forum?id=DqiggGDOmA}
}

@inproceedings{
xu2025deal,
title={{DEAL}: Diffusion Evolution Adversarial Learning for Sim-to-Real Transfer},
author={Wentao Xu and Huiqiao Fu and Haoyu Dong and Zhehao Zhou and Chunlin Chen},
booktitle={The Thirty-ninth Annual Conference on Neural Information Processing Systems},
year={2025},
url={https://openreview.net/forum?id=284GWLFtjU}
}

@misc{linuxinput2025,
  author       = {Linux Kernel Developers},
  title        = {Linux Kernel Documentation: Input Subsystem},
  year         = {2025},
  howpublished = {\url{https://docs.kernel.org/input/input.html}},
  note         = {Accessed: December 15, 2025}
}

@misc{zeromq2010,
  title        = {ZeroMQ},
  author       = {ZeroMQ Developers},
  year         = {2010},
  howpublished = {\url{https://zeromq.org/}},
}

@misc{ddsfoundation2025,
  author       = {DDS Foundation},
  title        = {What is DDS?},
  year         = {2025},
  howpublished = {\url{https://www.dds-foundation.org/what-is-dds-3/}},
}

@software{pink,
  title = {{Pink: Python inverse kinematics based on Pinocchio}},
  author = {Caron, Stéphane and De Mont-Marin, Yann and Budhiraja, Rohan and Bang, Seung Hyeon and Domrachev, Ivan and Nedelchev, Simeon and Du, Peter and Escande, Adrien and Vaillant, Joris and Wingo, Bruce},
  license = {Apache-2.0},
  url = {https://github.com/stephane-caron/pink},
  version = {3.5.0},
  year = {2025}
}

@inproceedings{Buss2004IntroductionTI,
  title={Introduction to Inverse Kinematics with Jacobian Transpose , Pseudoinverse and Damped Least Squares methods},
  author={Samuel R. Buss},
  year={2004},
  url={https://api.semanticscholar.org/CorpusID:5714366}
}

@misc{overwatch2017gdc,
  author       = {Ford, Timothy },
  title        = {Overwatch: Gameplay Architecture and Netcode},
  year         = {2017},
  howpublished = {\url{https://www.gdcvault.com/play/1024001/-Overwatch-Gameplay-Architecture-and}},
  note         = {Game Developers Conference (GDC) Vault}
}

@Article{         harris2020array,
 title         = {Array programming with {NumPy}},
 author        = {Charles R. Harris and K. Jarrod Millman and St{\'{e}}fan J.
                 van der Walt and Ralf Gommers and Pauli Virtanen and David
                 Cournapeau and Eric Wieser and Julian Taylor and Sebastian
                 Berg and Nathaniel J. Smith and Robert Kern and Matti Picus
                 and Stephan Hoyer and Marten H. van Kerkwijk and Matthew
                 Brett and Allan Haldane and Jaime Fern{\'{a}}ndez del
                 R{\'{i}}o and Mark Wiebe and Pearu Peterson and Pierre
                 G{\'{e}}rard-Marchant and Kevin Sheppard and Tyler Reddy and
                 Warren Weckesser and Hameer Abbasi and Christoph Gohlke and
                 Travis E. Oliphant},
 year          = {2020},
 month         = sep,
 journal       = {Nature},
 volume        = {585},
 number        = {7825},
 pages         = {357--362},
 doi           = {10.1038/s41586-020-2649-2},
 publisher     = {Springer Science and Business Media {LLC}},
 url           = {https://doi.org/10.1038/s41586-020-2649-2}
}

@inproceedings{qin2023anyteleop,
  title     = {AnyTeleop: A General Vision-Based Dexterous Robot Arm-Hand Teleoperation System},
  author    = {Qin, Yuzhe and Yang, Wei and Huang, Binghao and Van Wyk, Karl and Su, Hao and Wang, Xiaolong and Chao, Yu-Wei and Fox, Dieter},
  booktitle = {Robotics: Science and Systems},
  year      = {2023}
}

@misc{pybind11github,
  author       = {Jakob, Wenzel and Rhinelander, Jason and Moldovan, Dean},
  title        = {pybind11: Seamless Operability between C++11 and Python},
  year         = {2025},
  howpublished = {\url{https://github.com/pybind/pybind11}},
}

@ARTICLE{ros2compose,
  author={Macenski, Steve and Soragna, Alberto and Carroll, Michael and Ge, Zhenpeng},
  journal={IEEE Robotics and Automation Letters}, 
  title={Impact of ROS 2 Node Composition in Robotic Systems}, 
  year={2023},
  volume={8},
  number={7},
  pages={3996-4003},
  doi={10.1109/LRA.2023.3279614}}

@article{Chitta2017roscontrolAG,
  title={ros\_control: A generic and simple control framework for ROS},
  author={Sachin Chitta and Eitan Marder-Eppstein and Wim Meeussen and Vijay Pradeep and Adolfo Rodr{\'i}guez Tsouroukdissian and J. Aislinn Bohren and Dave Coleman and Bence Magyar and Gennaro Raiola and Mathias L{\"u}dtke and Enrique Fernandez Perdomo},
  journal={J. Open Source Softw.},
  year={2017},
  volume={2},
  pages={456},
  url={https://api.semanticscholar.org/CorpusID:53228319}
}

@article{Chitta2012MoveItT,
  title={MoveIt! [ROS Topics]},
  author={Sachin Chitta and Ioan Alexandru Sucan and Steve B. Cousins},
  journal={IEEE Robotics Autom. Mag.},
  year={2012},
  volume={19},
  pages={18-19},
  url={https://api.semanticscholar.org/CorpusID:206481888}
}

@INPROCEEDINGS{realsense,
  author={Keselman, Leonid and Woodfill, John Iselin and Grunnet-Jepsen, Anders and Bhowmik, Achintya},
  booktitle={2017 IEEE Conference on Computer Vision and Pattern Recognition Workshops (CVPRW)}, 
  title={Intel(R) RealSense(TM) Stereoscopic Depth Cameras}, 
  year={2017},
  volume={},
  number={},
  pages={1267-1276},
  doi={10.1109/CVPRW.2017.167}}

@article{Meier2015PX4AN,
  title={PX4: A node-based multithreaded open source robotics framework for deeply embedded platforms},
  author={Lorenz Meier and Dominik Honegger and Marc Pollefeys},
  journal={2015 IEEE International Conference on Robotics and Automation (ICRA)},
  year={2015},
  pages={6235-6240},
  url={https://api.semanticscholar.org/CorpusID:2440303}
}

@software{vuer,
  author = {Ge Yang},
  title = {{VUER}: An Event-Driven, Declarative Visualization Toolkit for GenAI and Robotics},
  version = {},
  publisher = {GitHub},
  url = {https://github.com/vuer-ai/vuer},
  year = {2025}
}

@article{Ansel2024PyTorch2F,
  title={PyTorch 2: Faster Machine Learning Through Dynamic Python Bytecode Transformation and Graph Compilation},
  author={Jason Ansel and Edward Yang and Horace He and Natalia Gimelshein and Animesh Jain and Michael Voznesensky and Bin Bao and Peter Bell and David Berard and Evgeni Burovski and Geeta Chauhan and Anjali Chourdia and Will Constable and Alban Desmaison and Zachary DeVito and Elias Ellison and Will Feng and Jiong Gong and Michael Gschwind and Brian Hirsh and Sherlock Huang and Kshiteej Kalambarkar and Laurent Kirsch and Michael Lazos and Mario Lezcano and Yanbo Liang and Jason Liang and Yinghai Lu and C. K. Luk and Bert Maher and Yunjie Pan and Christian Puhrsch and Matthias Reso and Mark-Albert Saroufim and Marcos Yukio Siraichi and Helen Suk and Shunting Zhang and Michael Suo and Phil Tillet and Xu Zhao and Eikan Wang and Keren Zhou and Richard Zou and Xiaodong Wang and Ajit Mathews and William Wen and Gregory Chanan and Peng Wu and Soumith Chintala},
  journal={Proceedings of the 29th ACM International Conference on Architectural Support for Programming Languages and Operating Systems, Volume 2},
  year={2024},
  url={https://api.semanticscholar.org/CorpusID:268794728}
}

@misc{newton2025github,
  author       = {NewtonPhysics},
  title        = {An open-source, GPU-accelerated physics simulation engine built upon NVIDIA Warp, specifically targeting roboticists and simulation researchers},
  year         = {2025},
  howpublished = {\url{https://github.com/newton-physics/newton}},
}

@misc{arxroboticsx2025github,
  author       = {ARXrobotics},
  title        = {ARXrobotics GitHub Repository},
  year         = {2025},
  howpublished = {\url{https://github.com/ARXroboticsX}},
}

@misc{agibot2023x2,
  author       = {AgiBot},
  title        = {AgiBot X2 Series},
  year         = {2023},
  howpublished = {\url{https://www.agibot.com/products/X2}},
}

@misc{rohand2025oymotion,
  author       = {OYMotion},
  title        = {ROHand},
  year         = {2025},
  howpublished = {\url{https://oymotion.com/en/product62/191}},
}

@article{Koenig2004DesignAU,
  title={Design and use paradigms for Gazebo, an open-source multi-robot simulator},
  author={Nathan P. Koenig and Andrew Howard},
  journal={2004 IEEE/RSJ International Conference on Intelligent Robots and Systems (IROS) (IEEE Cat. No.04CH37566)},
  year={2004},
  volume={3},
  pages={2149-2154 vol.3},
  url={https://api.semanticscholar.org/CorpusID:206941306}
}

@misc{nvidia2023isaacsim,
  author       = {NVIDIA},
  title        = {Isaac Sim - Robotics Simulation and Synthetic Data Generation},
  year         = {2023},
  howpublished = {\url{https://developer.nvidia.com/isaac-sim}},
}

@article{Michel2004CyberboticsLW,
  title={Cyberbotics Ltd. Webots{\texttrademark}: Professional Mobile Robot Simulation},
  author={Olivier Michel},
  journal={International Journal of Advanced Robotic Systems},
  year={2004},
  volume={1},
  url={https://api.semanticscholar.org/CorpusID:62708658}
}

@article{shacklett23madrona,
    title   = {An Extensible, Data-Oriented Architecture for
               High-Performance, Many-World Simulation},
    author  = {Brennan Shacklett and Luc Guy Rosenzweig and
               Zhiqiang Xie and Bidipta Sarkar and Andrew Szot and
               Erik Wijmans and Vladlen Koltun and Dhruv Batra and 
               Kayvon Fatahalian},
    journal = {ACM Trans. Graph.},
    volume  = {42},
    number  = {4},
    year    = {2023}
}

@article{Hughes1989WhyFP,
  title={Why Functional Programming Matters},
  author={John Hughes},
  journal={Comput. J.},
  year={1989},
  volume={32},
  pages={98-107},
  url={https://api.semanticscholar.org/CorpusID:16084483}
}

@article{ros2,
    author = {Steven Macenski and Tully Foote and Brian Gerkey and Chris Lalancette and William Woodall},
    title = {Robot Operating System 2: Design, architecture, and uses in the wild},
    journal = {Science Robotics},
    volume = {7},
    number = {66},
    pages = {eabm6074},
    year = {2022},
    doi = {10.1126/scirobotics.abm6074},
    URL = {https://www.science.org/doi/abs/10.1126/scirobotics.abm6074}
}

@inproceedings{xue2025unified,
  title={A Unified and General Humanoid Whole-Body Controller for Fine-Grained Locomotion}, 
  author={Xue, Yufei and Dong, Wentao and Liu, Minghuan and Zhang, Weinan and Pang, Jiangmiao},
  booktitle={Robotics: Science and Systems (RSS)},
  year={2025}
}

@inproceedings{Cheng2024OpenTeleVisionTW,
  title={Open-TeleVision: Teleoperation with Immersive Active Visual Feedback},
  author={Xuxin Cheng and Jialong Li and Shiqi Yang and Ge Yang and Xiaolong Wang},
  booktitle={Conference on Robot Learning},
  year={2024},
}

@Misc{fftai,
  title = {Fourier},
  author = {Fourier},
  howpublished = {\url{https://github.com/FFTAI}},
  year = 2022,
}

@misc{pnd,
  title = {PNDbotics Website},
  author = {PNDbotics},
  howpublished = {\url{https://pndbotics.com/humanoid}},
  year = 2025,
}

@misc{t1,
  title = {Booster Robotics Official Website},
  author = {Booster},
  howpublished = {\url{https://www.boosterobotics.com/robots/}},
  year = 2025,
}

@misc{oghr2,
  title = {loongOpen/OpenLoong-Hardware},
  author = {OpenLoong},
  howpublished = {\url{https://github.com/loongOpen/OpenLoong-Hardware}},
  year = 2025,
}

@misc{atom,
  title = {DOBOT Atom},
  author = {Dobot},
  howpublished = {\url{https://www.dobot-robots.com/products/humanoid-robots/atom.html}},
  year = 2025,
}

@inproceedings{he2024agile,
  author    = {He, Tairan and Zhang, Chong and Xiao, Wenli and He, Guanqi and Liu, Changliu and Shi, Guanya},
  title     = {Agile But Safe: Learning Collision-Free High-Speed Legged Locomotion},
  booktitle = {Robotics: Science and Systems (RSS)},
  year      = {2024},
}

@article{OpenAI2019SolvingRC,
  title={Solving Rubik's Cube with a Robot Hand},
  author={OpenAI and Ilge Akkaya and Marcin Andrychowicz and Maciek Chociej and Ma-teusz Litwin and Bob McGrew and Arthur Petron and Alex Paino and Matthias Plappert and Glenn Powell and Raphael Ribas and Jonas Schneider and Nikolas A. Tezak and Jerry Tworek and Peter Welinder and Lilian Weng and Qiming Yuan and Wojciech Zaremba and Lei M. Zhang},
  journal={ArXiv},
  year={2019},
  volume={abs/1910.07113},
  url={https://api.semanticscholar.org/CorpusID:204734323}
}

@misc{yin2025unitrackerlearninguniversalwholebody,
      title={UniTracker: Learning Universal Whole-Body Motion Tracker for Humanoid Robots}, 
      author={Kangning Yin and Weishuai Zeng and Ke Fan and Zirui Wang and Qiang Zhang and Zheng Tian and Jingbo Wang and Jiangmiao Pang and Weinan Zhang},
      year={2025},
      eprint={2507.07356},
      archivePrefix={arXiv},
      primaryClass={cs.RO},
      url={https://arxiv.org/abs/2507.07356}, 
}

@software{Genesis,
          author = {Genesis Authors},
          title = {Genesis: A Universal and Generative Physics Engine for Robotics and Beyond},
          month = {December},
          year = {2024},
          url = {https://github.com/Genesis-Embodied-AI/Genesis}
        }

@misc{mujoco_playground_2025,
  title = {MuJoCo Playground: An open-source framework for GPU-accelerated robot learning and sim-to-real transfer.},
  author = {Zakka, Kevin and Tabanpour, Baruch and Liao, Qiayuan and Haiderbhai, Mustafa and Holt, Samuel and Luo, Jing Yuan and Allshire, Arthur and Frey, Erik and Sreenath, Koushil and Kahrs, Lueder A. and Sferrazza, Carlo and Tassa, Yuval and Abbeel, Pieter},
  year = {2025},
  publisher = {GitHub},
  url = {https://github.com/google-deepmind/mujoco_playground}
}

@article{ji2024exbody2,
  title={ExBody2: Advanced Expressive Humanoid Whole-Body Control}, 
  author={Ji, Mazeyu and Peng, Xuanbin and Liu, Fangchen and Li, Jialong and Yang, Ge and Cheng, Xuxin and Wang, Xiaolong},
  journal={arXiv preprint arXiv:2412.13196},
  year={2024},
  }

@article{cheng2024express,
title={Expressive Whole-Body Control for Humanoid Robots},
author={Cheng, Xuxin and Ji, Yandong and Chen, Junming and Yang, Ruihan and Yang, Ge and Wang, Xiaolong},
journal={arXiv preprint arXiv:2402.16796},
year={2024}
}

@article{Lai2024WorldMP,
  title={World Model-based Perception for Visual Legged Locomotion},
  author={Hang Lai and Jiahang Cao and Jiafeng Xu and Hongtao Wu and Yunfeng Lin and Tao Kong and Yong Yu and Weinan Zhang},
  journal={ArXiv},
  year={2024},
  volume={abs/2409.16784},
  url={https://api.semanticscholar.org/CorpusID:272881101}
}

@inproceedings{ros,
  title={ROS: an open-source Robot Operating System},
  author={Quigley, Morgan and Conley, Ken and Gerkey, Brian and Faust, Josh and Foote, Tully and Leibs, Jeremy and Wheeler, Rob and Ng, Andrew Y and others},
  booktitle={ICRA workshop on open source software},
  volume={3},
  number={3.2},
  pages={5},
  year={2009},
  organization={Kobe, Japan}
}

@inproceedings{comparesim,
  title={Simulation tools for model-based robotics: Comparison of bullet, havok, mujoco, ode and physx},
  author={Erez, Tom and Tassa, Yuval and Todorov, Emanuel},
  booktitle={2015 IEEE international conference on robotics and automation (ICRA)},
  pages={4397--4404},
  year={2015},
  organization={IEEE}
}

@article{orbit,
  title={Orbit: A unified simulation framework for interactive robot learning environments},
  author={Mittal, Mayank and Yu, Calvin and Yu, Qinxi and Liu, Jingzhou and Rudin, Nikita and Hoeller, David and Yuan, Jia Lin and Singh, Ritvik and Guo, Yunrong and Mazhar, Hammad and others},
  journal={IEEE Robotics and Automation Letters},
  year={2023},
  publisher={IEEE}
}

@article{pybullet,
  title={Pybullet, a python module for physics simulation for games, robotics and machine learning},
  author={Coumans, Erwin and Bai, Yunfei},
  year={2016}
}

@article{isaacgym,
  title={Isaac Gym: High Performance GPU-Based Physics Simulation For Robot Learning},
  author={Viktor Makoviychuk and Lukasz Wawrzyniak and Yunrong Guo and Michelle Lu and Kier Storey and Miles Macklin and David Hoeller and N. Rudin and Arthur Allshire and Ankur Handa and Gavriel State},
  journal={ArXiv},
  year={2021},
  volume={abs/2108.10470}
}

@article{mujoco,
  title={MuJoCo: A physics engine for model-based control},
  author={Emanuel Todorov and Tom Erez and Yuval Tassa},
  journal={2012 IEEE/RSJ International Conference on Intelligent Robots and Systems},
  year={2012},
  pages={5026-5033}
}

@Misc{		  onnx,
  author	= {Bai, Junjie and Lu, Fang and Zhang, Ke and others},
  title		= {ONNX: Open Neural Network Exchange},
  year		= {2019},
  publisher	= {GitHub},
  journal	= {GitHub repository},
  howpublished	= {\url{https://github.com/onnx/onnx}},
  commit	= {94d238d96e3fb3a7ba34f03c284b9ad3516163be}
}

@InProceedings{	  pytorch,
  title		= {PyTorch: An Imperative Style, High-Performance Deep
		  Learning Library},
  author	= {Adam Paszke and Sam Gross and Francisco Massa and Adam
		  Lerer and James Bradbury and Gregory Chanan and Trevor
		  Killeen and Zeming Lin and Natalia Gimelshein and Luca
		  Antiga and Alban Desmaison and Andreas K{\"o}pf and Edward
		  Yang and Zach DeVito and Martin Raison and Alykhan Tejani
		  and Sasank Chilamkurthy and Benoit Steiner and Lu Fang and
		  Junjie Bai and Soumith Chintala},
  booktitle	= {NeurIPS},
  year		= {2019}
}

@article{campanaro2022,
  title =        {Learning and deploying robust locomotion policies with minimal
                  dynamics randomization},
  author =       {Campanaro, Luigi and Gangapurwala, Siddhant and Merkt,
                  Wolfgang and Havoutis, Ioannis},
  journal =      {arXiv preprint arXiv:2209.12878},
  year =         2022
}

@article{smith2022,
  title =        {A walk in the park: Learning to walk in 20 minutes with
                  model-free reinforcement learning},
  author =       {Smith, Laura and Kostrikov, Ilya and Levine, Sergey},
  journal =      {arXiv preprint arXiv:2208.07860},
  year =         2022
}
\bibliographystyle{rlc}

\end{document}